\documentclass[10pt,twocolumn,letterpaper]{article}

\usepackage{configs/ijcb}
\usepackage{times}
\usepackage{epsfig}
\usepackage{graphicx}
\usepackage{amsmath}
\usepackage{amssymb}
\usepackage{multirow}
\usepackage{graphicx}
\usepackage{pifont}
\newcommand{\cmark}{\ding{51}}%
\newcommand{\xmark}{\ding{55}}%

\ijcbfinalcopy % *** Uncomment this line for the final submission

\ifijcbfinal\fi
\begin{document}

%%%%%%%%% TITLE
\title{LADIMO: Face Morph Generation through Biometric Template Inversion with Latent Diffusion}

\author{Marcel Grimmer \\
NTNU\thanks{Norwegian University of Science and Technology}  \qquad h\_da\thanks{Hochschule Darmstadt}\\
Gj{\o}vik, Norway \qquad Darmstadt, Germany\\
{\tt\small marceg@ntnu.no}
\and 
Christoph Busch\\
NTNU\footnotemark[1]  \qquad h\_da\footnotemark[2]\\
Gj{\o}vik, Norway \qquad Darmstadt, Germany\\
{\tt\small christoph.busch@ntnu.no}
}

\maketitle
\thispagestyle{empty}

\begin{abstract}
Face morphing attacks pose a severe security threat to face recognition systems, enabling the morphed face image to be verified against multiple identities. To detect such manipulated images, the development of new face morphing methods becomes essential to increase the diversity of training datasets used for face morph detection. In this study, we present a representation-level face morphing approach, namely \textit{LADIMO}, that performs morphing on two face recognition embeddings. Specifically, we train a \textit{Latent Diffusion Model} to invert a biometric template - thus reconstructing the face image from an FRS latent representation. Our subsequent vulnerability analysis demonstrates the high morph attack potential in comparison to MIPGAN-II, an established GAN-based face morphing approach. Finally, we exploit the stochastic LADMIO model design in combination with our identity conditioning mechanism to create unlimited morphing attacks from a single face morph image pair. We show that each face morph variant has an individual attack success rate, enabling us to maximize the morph attack potential by applying a simple re-sampling strategy. We will publish our code and pre-trained models upon the acceptance of this paper.       
\end{abstract}
\section{Introduction}

Nowadays, face recognition systems (FRS) find application in critical areas such as border control~\cite{EU-Regulation-EES-InternalDocument-2017} and forensics~\cite{EU-Regulation-2017-2226-on-EES-171130}\cite{EU-ImplementingDecision-2019-329-on-EES-SampleQuality-190225} due to their high accuracy, uniqueness, and non-intrusiveness. However, despite these advantages, one of the main challenges remains the distinction between intra-identity and inter-identity variability necessary to discern mated from non-mated comparison trials. In this context, the security of FRS depends on the decision threshold at which a presented probe image is verified successfully against the stored reference.

By exploiting low-security or misconfigured decision thresholds, face morph attacks aim to generate facial images with identity characteristics from two or multiple contributing subjects. Consequently, when successful, the morphed face image can be verified against each contributing identity. This poses a significant security threat, especially considering that in many countries, passport application and issuance still rely on printed self-delivered photos. 

\subsection{Face Morph Generation}

\begin{figure}
\label{fig:ladimo-intro}
\centering
\includegraphics[width=0.99\linewidth]{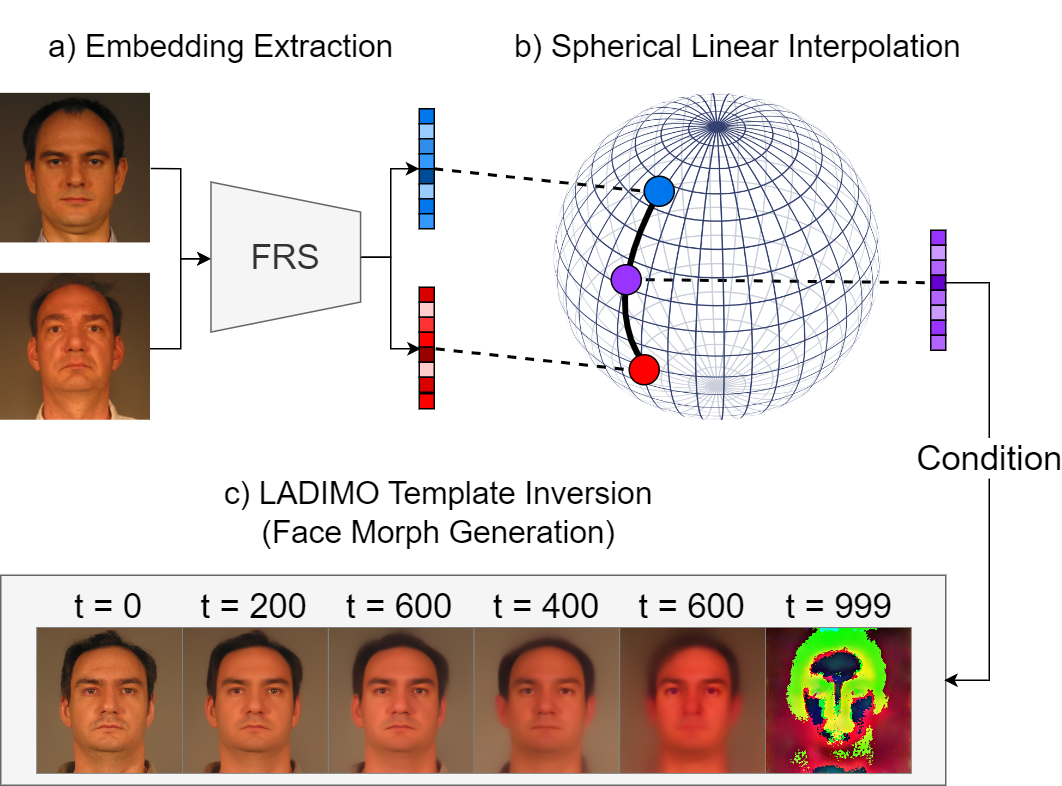}
\caption{Conceptual overview on LADIMO face morph generation, including a) the biometric template encoding~\cite{meng2021magface}, b) spherical linear interpolation~\cite{shoemake1985animating}, and c) LDM-based biometric template inversion results after t denoising steps.}
\end{figure}

To mitigate face morph attacks, training human examiners and machine learning-based models to detect manipulated facial images becomes crucial. Addressing this challenge, developing a diverse set of face morph approaches supports the detection generalizability, as different generative models leave individual patterns or artefacts~\cite{ibsen2023multi} and hence a larger diversity in the training material can lead to better detection accuracy for unseen methods. Hence, current research~\cite{venkatesh2021face} highlights the necessity of generating diverse face morph datasets to facilitate robust morph detection.  

Generally, face morph generation techniques can operate on the \textit{representational-level}~\cite{Zhang-MIPGAN-TBIOM-2021}\cite{Colbois-OptimalFaceMorphs-IJCB-2023}\cite{Damer-MorDiff-IWBF-2023} or \textit{image-level}~\cite{Ferrara-ImageLevelMorphing-p1-TIFS-2017}\cite{Raghavendra-ImageLevelMorphing-p2-IJCB-2017}. Image-level face morphing typically involves computing triangulated landmarks of contributing images, followed by affine transformations, warping, and alpha blending to create a morphed face image~\cite{raghavendra2017face}.  While image-based face morphing exhibits strong identity preservation and poses significant security risks for FRS, it is also prone to visible artefacts that necessitate manual post-editing, hence limiting its scalability in terms of generating large-scale training databases that can eventually lead to more robust morph detection mechanisms. 

Instead, generating face morphs at the representational level involves encoding contributing facial images into semantically meaningful feature vectors, also known as \textit{latent representations} or \textit{latent embeddings}. Subsequently, the semantic features are morphed by interpolating between them, after which the resulting latent embedding is passed to a decoder to generate the corresponding morphed face image. Typically, representation-level face morphing operates within the latent space of generative models, such as \textit{Generative Adversarial Networks} (GAN)~\cite{Zhang-MIPGAN-TBIOM-2021} or \textit{Diffusion Autoencoders}~\cite{Damer-MorDiff-IWBF-2023}.

\subsection{Contribution}

We harness the high visual fidelity and conditioning ability of \textit{Latent Diffusion Models} (LDM)~\cite{Rombach-LDM-CVPR-2022}, applying these properties to the domain of face morph generation. Notably, unlike Damer et al.\cite{Damer-MorDiff-IWBF-2023} and Blasingame et al.\cite{blasingame2024leveraging}, who utilize \textit{Diffusion Autoencoders}~\cite{preechakul2022diffusion} with general-purpose semantic representations, our focus is on directly inverting biometric templates to maximize the \textit{Morph Attack Potential} (MAP)~\cite{ISO-IEC-20059}. This idea is similar to Colbois et al.\cite{Colbois-OptimalFaceMorphs-IJCB-2023}, who translate biometric templates to the intermediate latent space of StyleGAN 3~\cite{karras2021alias} through re-training its mapping network while keeping the synthesis module fixed during training. While GAN-based face morphing techniques excel at generating strong morph attacks, the reconstructed images often suffer from unnatural skin textures when manipulating latent representation deviating from the average face image of the training dataset (see Figure~\ref{fig:morph-visuals}). To address this challenge, we propose a latent diffusion-based face morphing method (LADIMO) that generates photorealistic face image reconstructions.

In summary, our contributions can be outlined as follows:

\begin{itemize}
    \item We introduce LADIMO, a LDM-based face morphing model designed to invert biometric MagFace~\cite{meng2021magface} templates, enabling representational-level morphing through \textit{spherical linear interpolation}~\cite{shoemake1985animating}.
    \item We conduct an extensive vulnerability analysis, assessing the MAP on four state-of-the-art FRS in comparison to a traditional GAN-based face morphing method (MIPGAN-II~\cite{Zhang-MIPGAN-TBIOM-2021}).
    \item We leverage the architectural design of LADIMO to introduce the concept of \textit{Stochastic Morph Variation}, allowing us to maximize the MAP by treating morphed images as random variables.
\end{itemize}

This paper is structured as follows: First, Section~\ref{sec:proposedMethod} introduces the fundamental concept of LADIMO and how it is applied to inverting biometric templates for generating face morph attacks. Next, Section~\ref{sec:experimentalSettings} provides details about the datasets and morph pair selection strategy. In Section~\ref{sec:experimentalResults}, we present and discuss our experimental findings, with a particular focus on visual inspection, FRS vulnerability analysis, and the optimization of morph attacks through stochastic morph variation. Finally, Section~\ref{sec:summary} summarises the main limitations, findings, and directions for future work. 
\section{Proposed LADIMO Model}
\label{sec:proposedMethod}

\subsection{Biometric Template Inversion}

\begin{figure}
\centering
\includegraphics[width=0.99\linewidth]{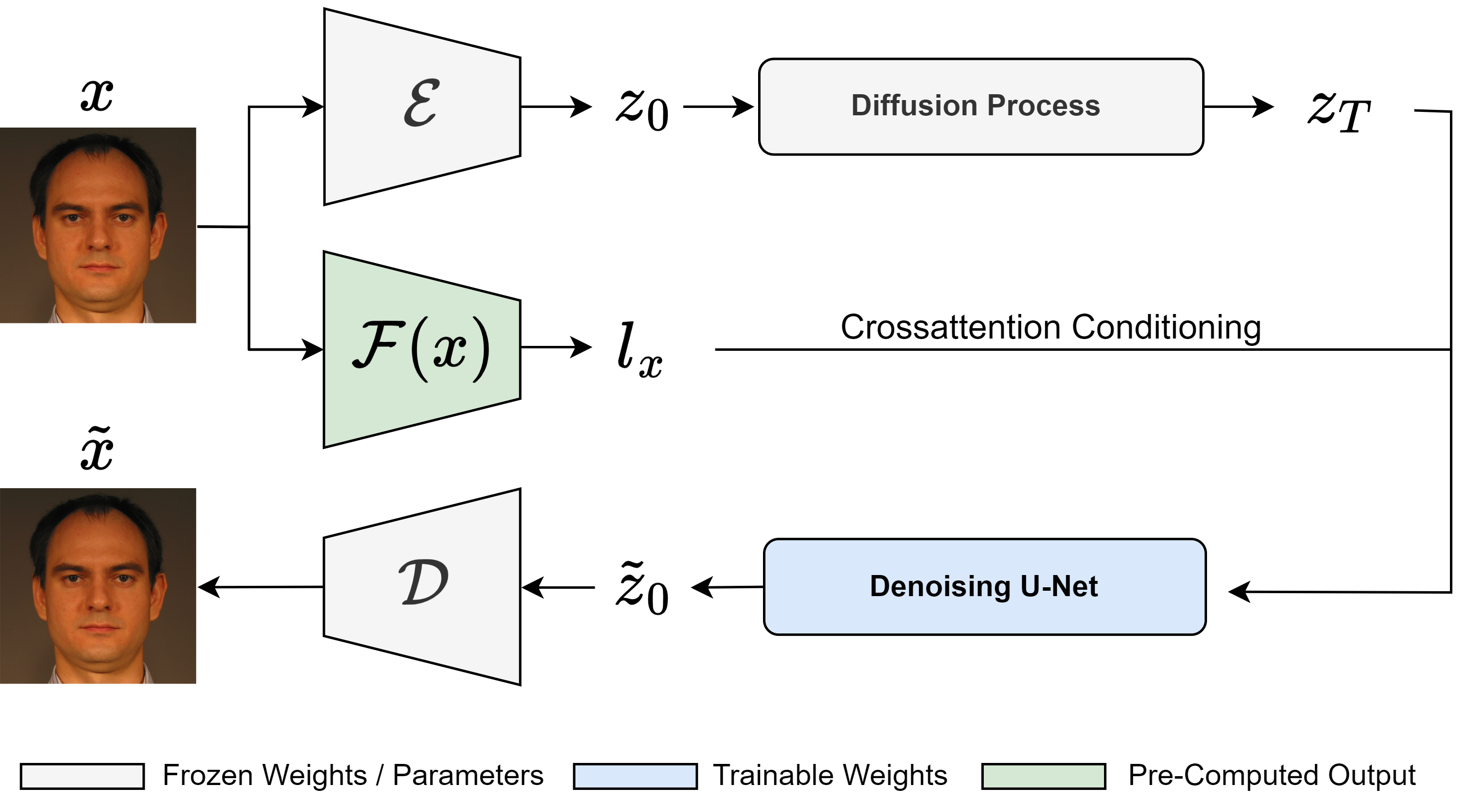}
\caption{LADIMO training architecture overview, adopting the default hyperparameter settings from~\cite{Rombach-LDM-CVPR-2022} and utilizing their pre-trained perceptual encoder and decoder. We customize the conditioning mechanism for biometric template inversion, learning to reconstruct face images from their corresponding MagFace~\cite{meng2021magface} embeddings.}
\label{fig:ladimo-architecture-overview}
\end{figure}

Given a facial image $x$ and a FRS feature extractor $\mathcal{F}(.)$, we denote a latent representation (\textit{i.e.}, biometric template) as $l = \mathcal{F}(x)$. The main goal of biometric template inversion is to find an inverse function $\mathcal{F}^{-1}(.)$ such that $\Tilde{x} = \mathcal{F}^{-1}(l)$ reconstructs $x$ from its biometric template $l$. In this work, we train a conditioned LDM to approximate $\mathcal{F}^{-1}(.)$.

As shown in Figure~\ref{fig:ladimo-architecture-overview}, we adhere to the default training setup of \cite{Rombach-LDM-CVPR-2022}, wherein the encoder-decoder networks employed for perceptual compression\footnote{Setup: VQ Regularization, f=4, Z=8192, d=3} are frozen, and only the \textit{time-conditional UNet}~\cite{ronneberger2015u} for the denoising process is optimized. Initially, the perceptual encoder $\mathcal{E}$ encodes $x$ to $z_{0}$ from the \textit{pixel domain} to the \textit{latent domain}. After $T$ diffusion steps, the reverse diffusion from Gaussian noise $z_T$ back to the initial stage $z_0$ is conditioned by the MagFace~\cite{meng2021magface} latent embedding of $x$ through \textit{Cross-Attention}~\cite{vaswani2017attention}. This conditioning mechanism ensures that the denoising UNet is guided with a generalized understanding of the identity information encoded in the biometric template. Thus, during inference, the $\mathrm{LDM}$ can be utilized to invert latent representations by randomly drawing $z_T$ from a standard normal distribution $\mathcal{N}(0,1)$. The combination of random sampling and deterministic identity conditions allows for the generation of unlimited facial image variations with fixed identity characteristics. In our study, we refer to this concept as \textit{stochastic morph variation}, which is analysed in-depth in Section~\ref{sec:stochastic-morph-variation}.

LADIMO is developed in a two-stage training process: Initially, we train our LDM for 80 epochs on the FFHQ dataset~\cite{karras2019style}, consisting of 70,000 facial images acquired in uncontrolled environments, covering a wide diversity of facial attributes and demographics. Finally, we fine-tune our model for an additional 20 epochs using our FRGCv2~\cite{phillips2005overview} reference subset (see Section~\ref{sec:datasets}), tailoring the biometric template inversion to our face morphing task and maximizing the MAP. LADIMO is trained on a virtual NVIDIA A100 graphics card with 20GB RAM and a batch size of 8, using an AdamW optimizer.

\subsection{Face Morph Generation}

Once our LDM is trained, we generate our face morph attacks by reconstructing a facial image with identity characteristics from two distinct subjects. Following~\cite{shoemake1985animating}, we apply \textit{Spherical linear interpolation} (SLERP) with $\gamma = 0.5$ between the latent representations $l_1$ and $l_2$ of two contributing subjects:

\begin{equation}
    \text{SLERP}(l_1, l_2, \gamma) = \frac{\sin{(1-\gamma)\theta}}{\sin{\theta}}\cdot l_1 + \frac{\sin{\gamma\theta}}{\sin{\theta}}\cdot l_2
\end{equation}

where 

\begin{equation}
    \theta = \frac{\arccos{l_1\cdot l_2}}{||l_1||\cdot||l_2||}
\end{equation}

Hence, LADIMO can be classified as a representation-level morphing approach that operates directly on the FRS's biometric templates. Given the conditioned LDM, we can invert the interpolated latent representation $l_M$ to guide the denoising process starting from $z_T \sim N(0,1)$, drawn from a Gaussian distribution.

Operating directly on the biometric templates ensures the morph of identity features that contribute most to the verification decision. Theoretically, this poses the highest MAP as the face morph maximizes the cosine similarity between both contributing subjects, which is the decision criterion of FRS to discern mated from non-mated samples. Our experiments investigate how MagFace-specific face morph attacks translate to FRS independent from the training process, replicating attack scenarios where the attacker has no prior knowledge about internal systems. Thus, Section~\ref{sec:vulnerability-analysis} analyses the generalizability of our attacks based on their MAP~\cite{ISO-IEC-20059} matrices, demonstrating that LADIMO morph attacks can deceive FRS beyond those utilized during training.

\section{Evaluation Setup}
\label{sec:experimentalSettings}

\subsection{Datasets}
\label{sec:datasets}

\begin{figure}
\centering
\setlength{\tabcolsep}{1pt}
\begin{tabular}{ccc}

\includegraphics[width=.2\linewidth]{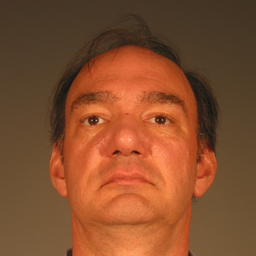} & \includegraphics[width=.2\linewidth]{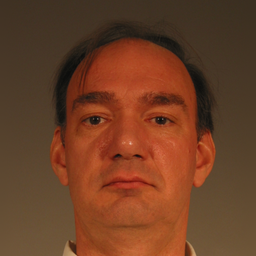} & \includegraphics[width=.2\linewidth]{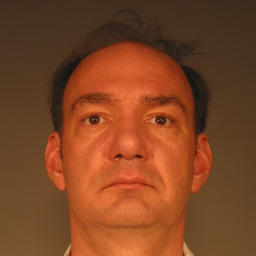} \\

\includegraphics[width=.2\linewidth]{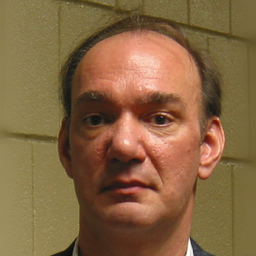} & \includegraphics[width=.2\linewidth]{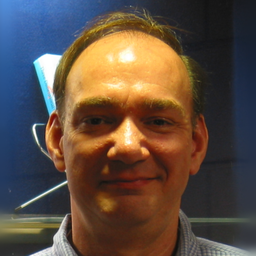} & \includegraphics[width=.2\linewidth]{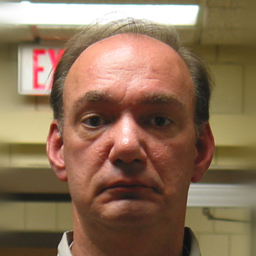} \\

\end{tabular}
\caption{Example reference (top) and probe (bottom) samples from our FRGCv2~\cite{phillips2005overview} subset.}
\label{fig:db-examples}
\end{figure}
\begin{table}[]
\caption{Overview of our reference and probe subsets.}
\resizebox{\linewidth}{!}{%
\begin{tabular}{|l|c|c|c|}
\hline
\textbf{Subset} & \textbf{Images} & \multicolumn{1}{l|}{\textbf{IDs}} & \multicolumn{1}{l|}{\textbf{Controlled Environment?}} \\ \hline
FRGCv2-Reference  & 1,440           & \multirow{2}{*}{325}              & \cmark                                 \\
FRGCv2-Probe      & 975             &                                   & \xmark                                 \\ \hline
\end{tabular}%
}
\label{tab:datasets}
\end{table}

We perform our face morph evaluation on the FRGCv2 dataset~\cite{phillips2005overview}. From the original dataset, we create one subset with controlled facial images (references) and another subset with semi-controlled facial images (probes). This data partitioning imitates the typical passport verification process at automatic border check (ABC) gates, where the machine-readable travel document stores a reference facial image with a controlled background, illumination, and facial expression. In contrast, the probe image captured directly at the ABC gate exposes the FRS to more variation in facial and environmental attributes. We present example facial images from the reference and probe subsets in Figure~\ref{fig:db-examples} along with additional information in Table~\ref{tab:datasets}.

\subsection{Morph Pair Selection}

The face morph generation is conducted only on the controlled reference facial images, as those would be presented in the passport application procedure. In this context, it is crucial to select the face morph pairs based on the similarity of the contributing subjects. There is a high positive correlation between the morph attack success rate and the initial similarity of the contributing subjects. At the representational level, a similarity-based pair selection interpolates between latent representations, which are already located near each other in the latent space - yielding face morphs with low spatial distances to their origins. In our experiments, the pair selection is based on non-mated similarity scores computed with ArcFace~\cite{Deng-ArcFace-CVPR-2019}.
\section{Experimental Results}
\label{sec:experimentalResults}

\subsection{Visual Analysis}
\begin{figure}
\centering
\setlength{\tabcolsep}{1pt}
\begin{tabular}{cccc}
Subject 1 & MIPGAN-II & LADIMO & Subject 2 \\
\includegraphics[width=.24\linewidth]{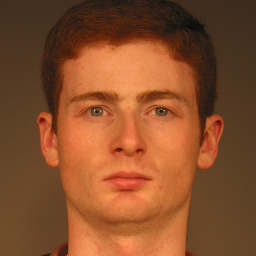} & \includegraphics[width=.24\linewidth]{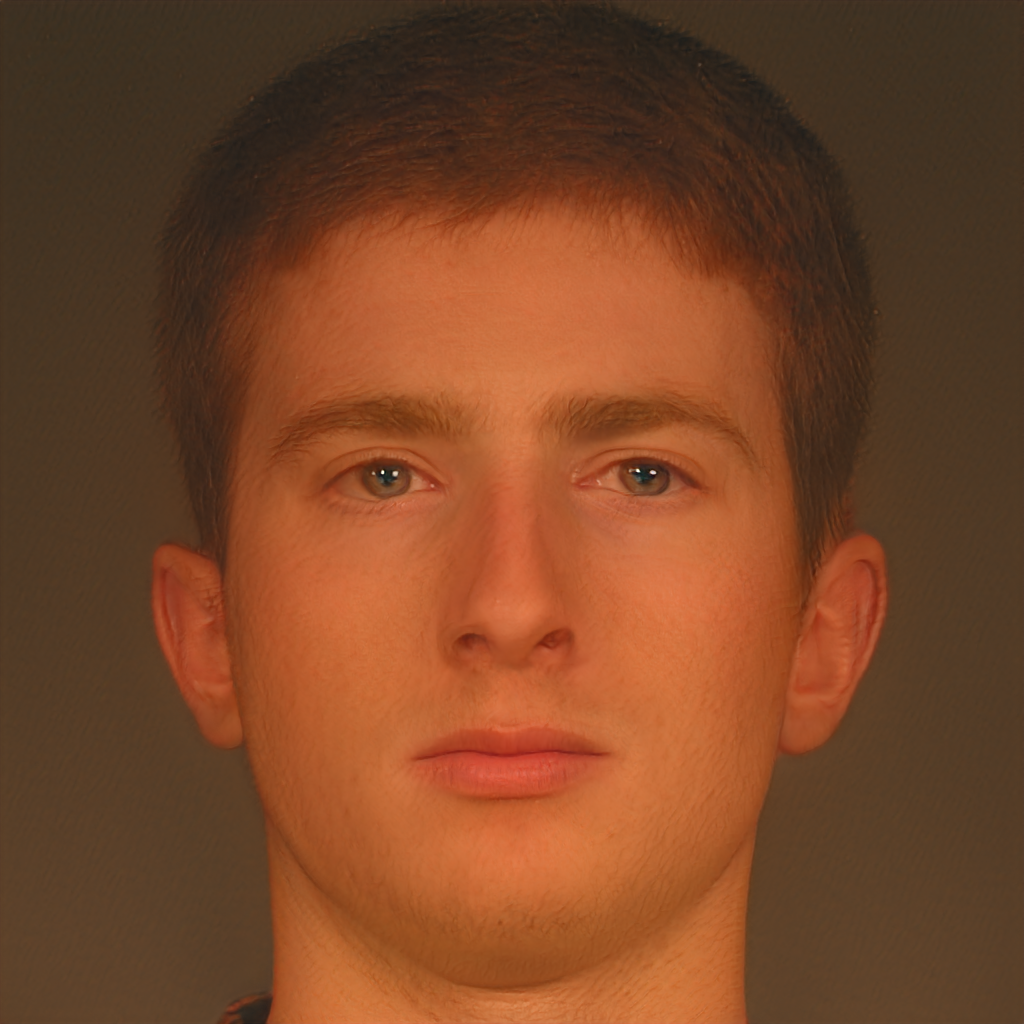} & \includegraphics[width=.24\linewidth]{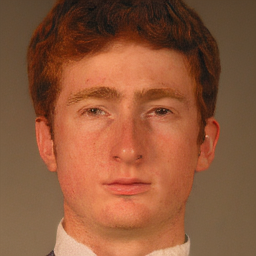} & \includegraphics[width=.24\linewidth]{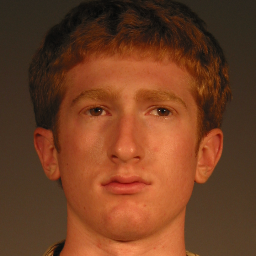} \\

\includegraphics[width=.24\linewidth]{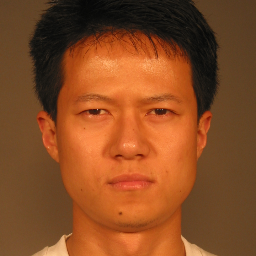} & \includegraphics[width=.24\linewidth]{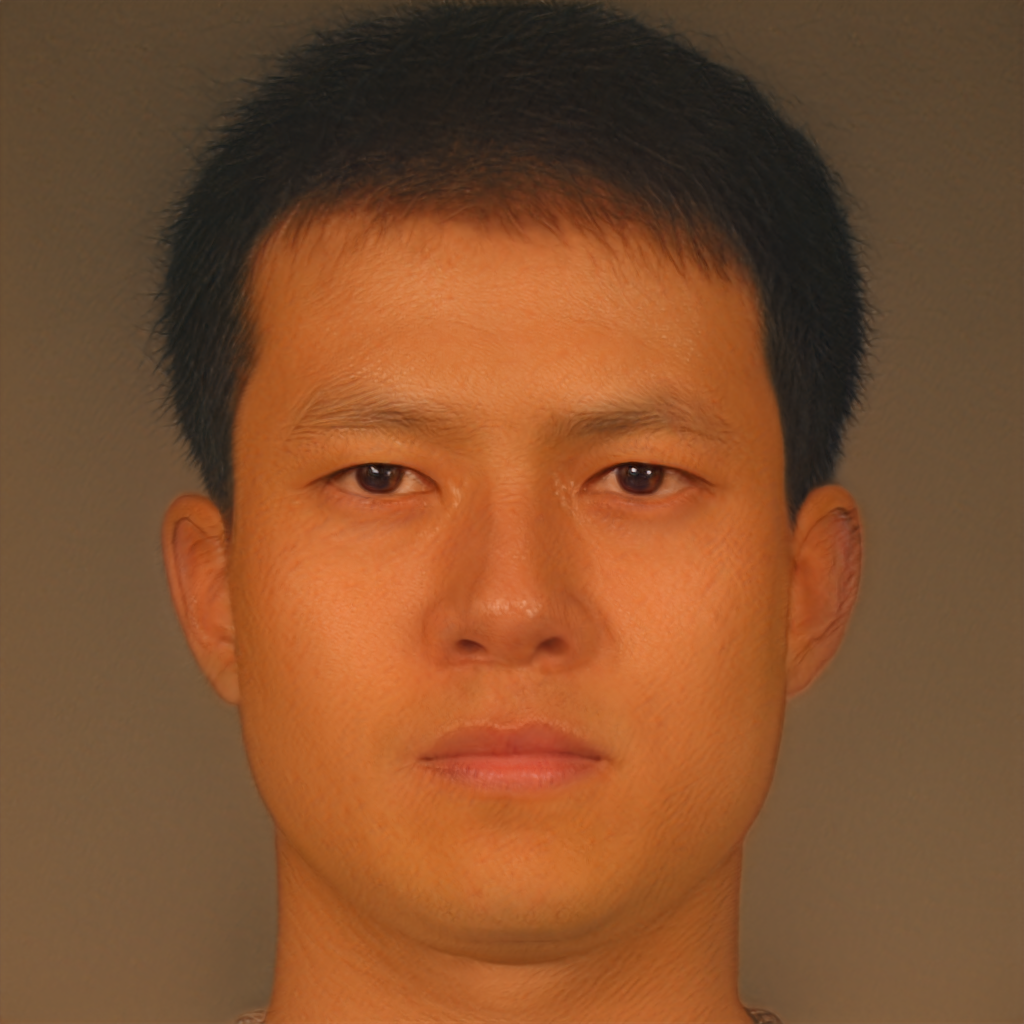} & \includegraphics[width=.24\linewidth]{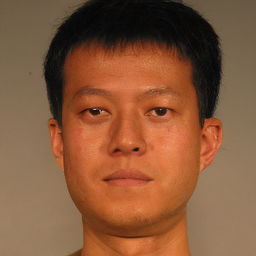} & \includegraphics[width=.24\linewidth]{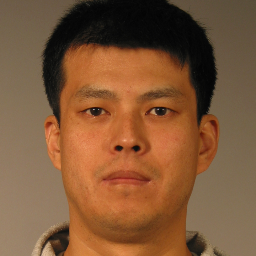} \\

\includegraphics[width=.24\linewidth]{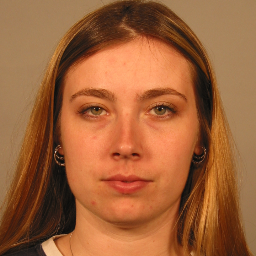} & \includegraphics[width=.24\linewidth]{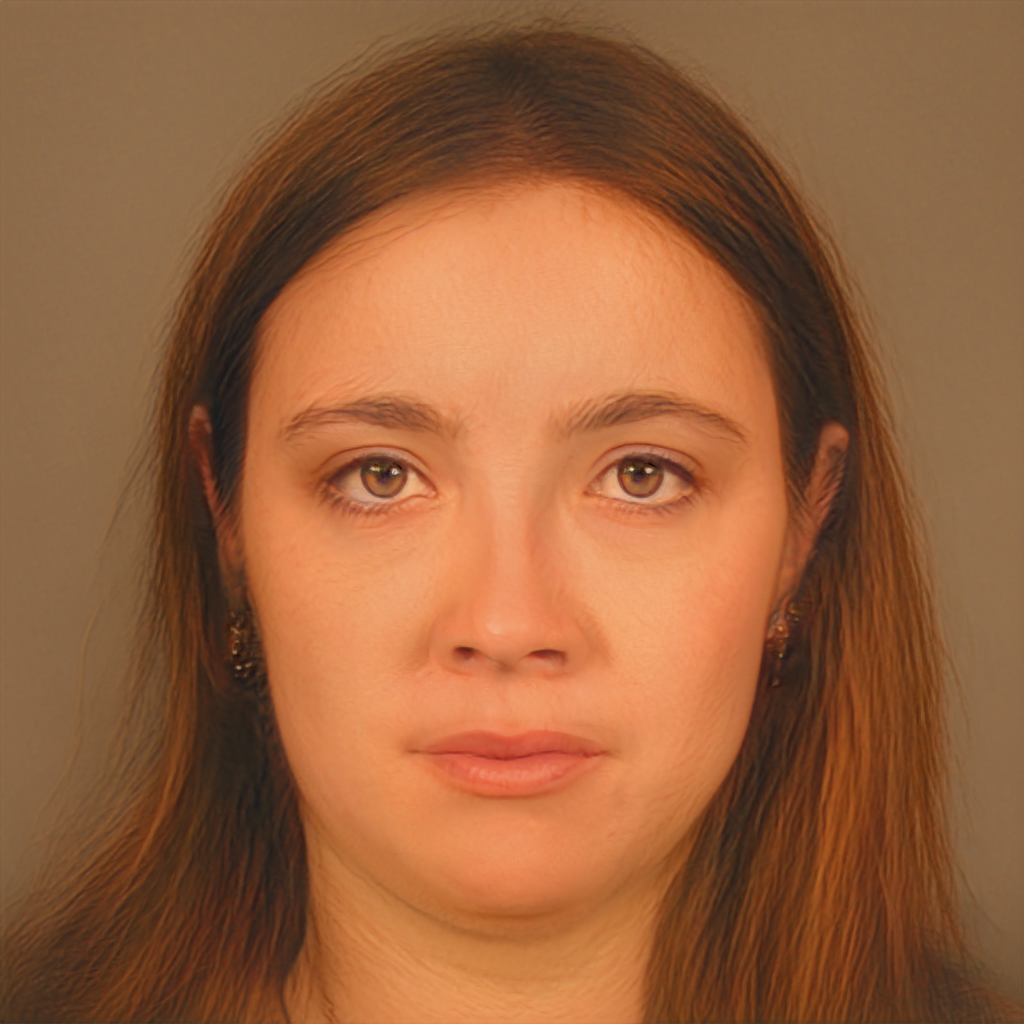} & \includegraphics[width=.24\linewidth]{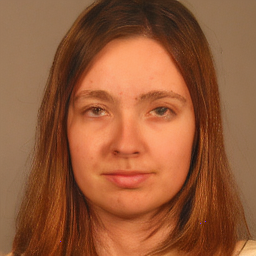} & \includegraphics[width=.24\linewidth]{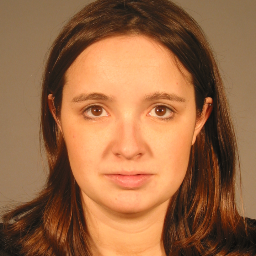} \\
\end{tabular}
\caption{Example face morph images generated with LADIMO in comparison to MIPGAN-II~\cite{Zhang-MIPGAN-TBIOM-2021}.}
\label{fig:morph-visuals}
\end{figure}

\begin{figure*}
\centering
\includegraphics[width=0.9\linewidth]{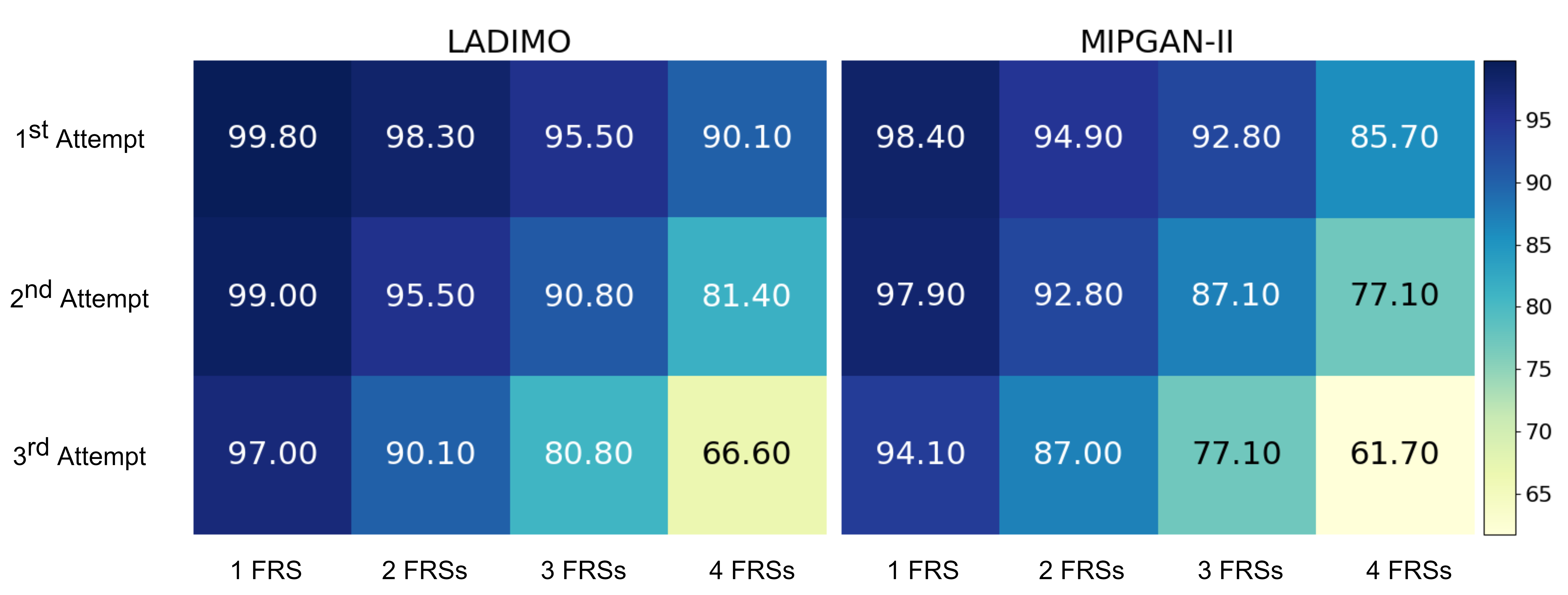}
\caption{LADIMO Morph Attack Potential~\cite{ISO-IEC-20059} in comparison to MIPGAN-II~\cite{Zhang-MIPGAN-TBIOM-2021}, assessed with three verification attempts and four FRSs: \cite{Deng-ArcFace-CVPR-2019}\cite{Kim-AdaFace-CVPR-2022}\cite{Huang-Curricularface-CVPR-2020}\cite{meng2021magface} with a fixed FMR at 0.1\%.}
\label{fig:ladimo-mipgan-map}
\end{figure*}

In Figure~\ref{fig:morph-visuals}, we showcase example face morphs representative of the image quality achieved with our proposed LADIMO template inversion approach. Furthermore, we conduct a comparative analysis of LADIMO and MIPGAN-II~\cite{Zhang-MIPGAN-TBIOM-2021} to study the differences between diffusion and GAN-based face morphing. A notable dissimilarity is evident in the skin texture comparison: LADIMO generates highly realistic skin textures that are challenging to distinguish from real data by human perception. Conversely, MIPGAN-II face morphs display over-smoothed skin surfaces, contributing to an unnatural appearance. This observation is pivotal, considering the necessity for face morph attacks to deceive human operators during enrolment in the official passport issuance.

We attribute the over-smoothed skin texture in GAN-generated facial images to information losses during the inversion of real facial images into the latent space, known as the \textit{information bottleneck theory}~\cite{tishby2015deep}. Additionally, GAN-based image synthesis encounters challenges when editing facial images that significantly deviate from the average facial image observed during training. Typically, editability performs best when sampling latent representations close to the average of the prior GAN distribution. To address these issues, we develop LADIMO based on LDM~\cite{Rombach-LDM-CVPR-2022}, employing perceptual compression while retaining semantic information. Consequently, the information-rich LDM latent space enables the reconstruction of facial images with higher fidelity and better identity preservation rates.

\subsection{Vulnerability Analysis}
\label{sec:vulnerability-analysis}

As the main objective of face morph attacks is to deceive both humans and FRS, we assess the MAP~\cite{ferrara2022morphing} to evaluate the attack success rates of LADIMO and MIPGAN-II~\cite{Zhang-MIPGAN-TBIOM-2021} across multiple FRS and verification attempts (see Figure~\ref{fig:ladimo-mipgan-map}). Specifically, we follow the definition of ISO/IEC CD 20059~\cite{ISO-IEC-20059} that specifies MAP as a \textit{measure of the capability of a morphing attack to deceive one or more biometric recognition systems using multiple recognition attempts}. Accordingly, each matrix cell MAP[r,c] expresses the proportion of morphed images successfully reaching a match decision against at least r verification attempts for each contributing subject by at least c of the evaluated FRS~\cite{Deng-ArcFace-CVPR-2019}\cite{Kim-AdaFace-CVPR-2022}\cite{Huang-Curricularface-CVPR-2020}\cite{meng2021magface}. 

In Figure~\ref{fig:ladimo-mipgan-map}, we measure the MAP based on a fixed FR threshold at a false match rate (FMR) of 0.1\%, aligned with the FRONTEX best practice guidelines~\cite{FRONTEX-BorderControl-BestPractices-InternalDocument-2015} and determined based on over 1 million non-mated comparison scores computed on FRGCv2~\cite{phillips2005overview}. The MAP matrices indicate that LADIMO consistently outperforms MIPGAN-II, with a widening gap as the number of verification attempts and FRS increases. Notably, even though encountering FRS not seen during training, LADIMO maintains a MAP above $66\%$, demonstrating its effectiveness and invariance to the evaluated FRS. 

\begin{figure}
\centering
\includegraphics[width=0.99\linewidth]{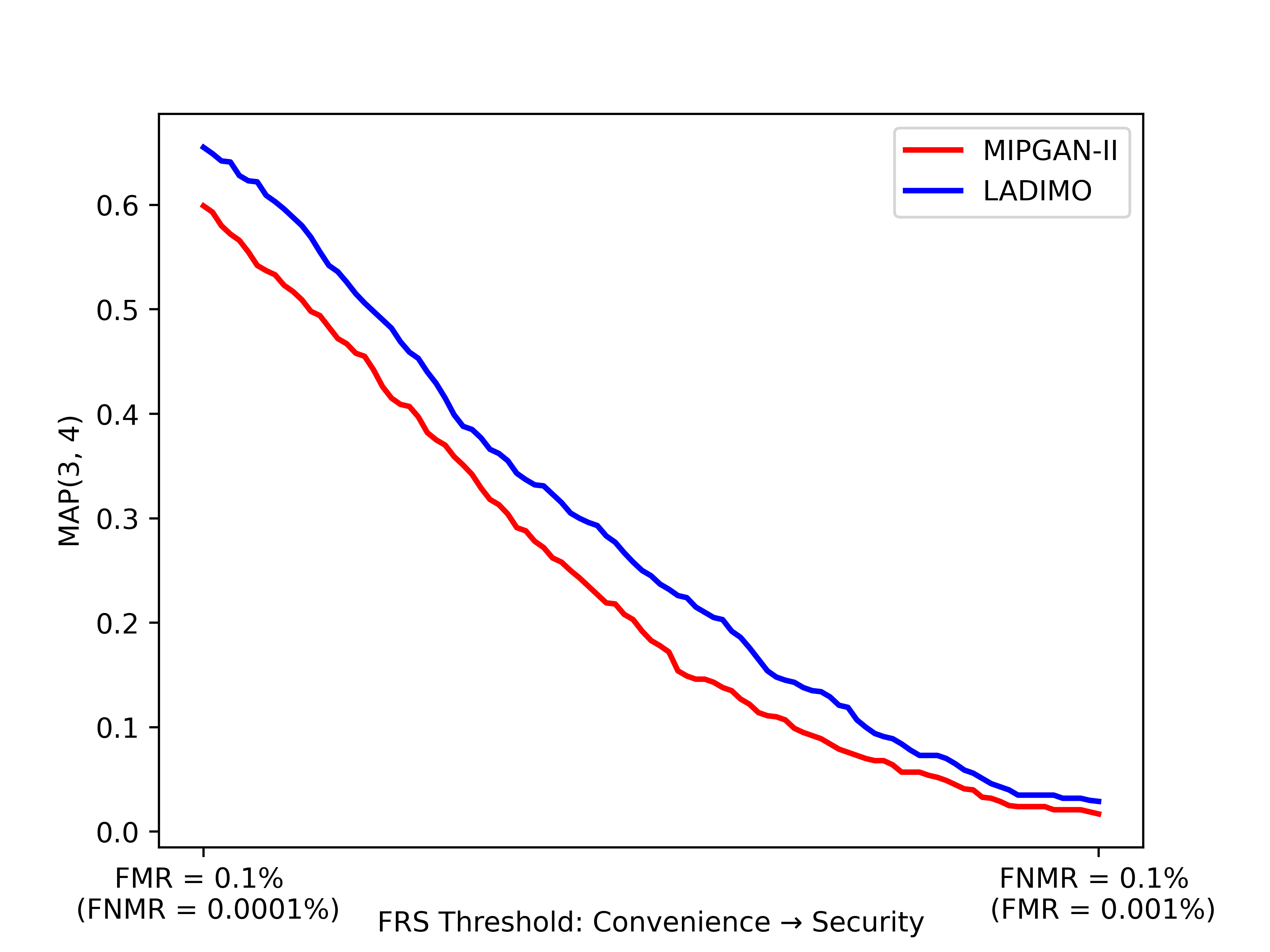}
\caption{Vulnerability Analysis showing how MAP(3, 4) changes across face recognition match decision thresholds relatively varied from a convenient-oriented level to a security-oriented level. FMRs and FNMRs are reported as averaged values over the four evaluated FRSs.}
\label{fig:FRS-MAP-threshold-analysis}
\end{figure}

The subsequent analysis, presented in Figure~\ref{fig:FRS-MAP-threshold-analysis}, reveals how MAP[3, 4] changes as the operational face recognition threshold shifts from a convenience-focused level to a more security-focused level. It becomes visible that the robustness of FRS towards face morph attacks significantly depends on the match decision threshold. Through adjusting the threshold, a higher proportion of face morph attacks are rejected because they cannot reach the required similarity to achieve a match decision. Besides following external best practice guidelines, this highlights the importance of tuning hyperparameters in biometric systems to fit the target application and reach the desired trade-off between convenience and security.  

\subsection{Stochastic Morph Variation}
\label{sec:stochastic-morph-variation}

In this section, we conduct an analysis of the stochastic face morph variation that allows the generation of unlimited face morphs with fixed contributing facial images. This sampling strategy is rooted in reconstructing a facial image from $z_T$ drawn from a standard normal distribution while fixing the morphed MagFace embedding $l_M$. Consequently, the image reconstruction can be treated as a stochastic process that can be exploited to optimize the morph attack. Specifically, our objective is to assess whether randomly generated face morphs with fixed contributing identities exhibit consistent attack potential or if re-sampling can be used to maximize the attack success rate.  

\begin{figure}[h!]
\centering
\includegraphics[width=0.99\linewidth]{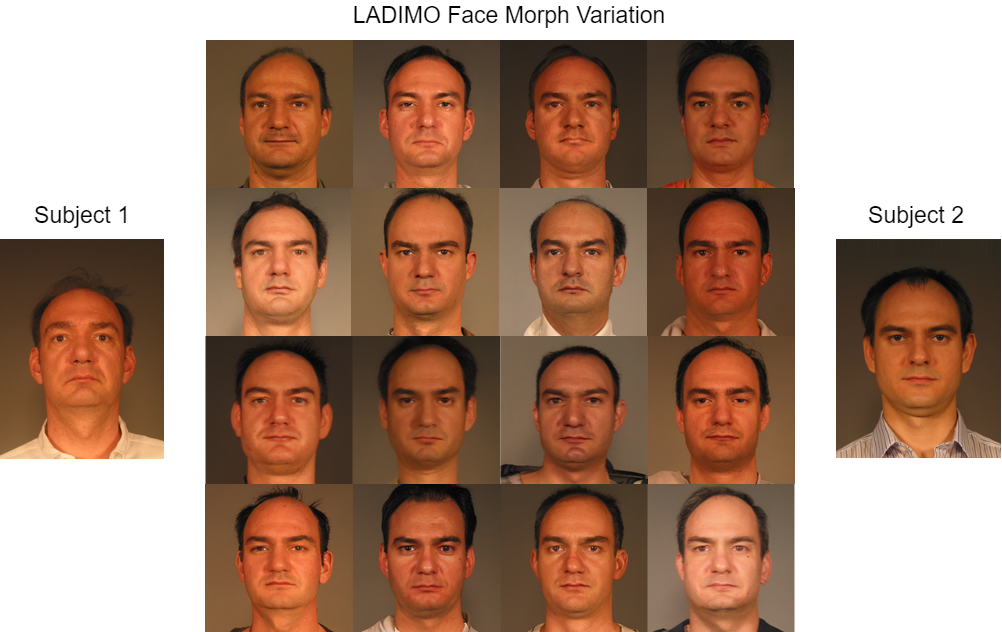}
\caption{Random selection of 16 morph variants generated with LADIMO derived from a fixed image pair.}
\label{fig:ladimo-single-id-morphs}
\end{figure}

A demonstration of the stochastic morph variation is shown in Figure~\ref{fig:ladimo-single-id-morphs}, presenting 16 randomly generated face morphs, where $z_T$ was drawn from a Gaussian distribution while keeping the morphed MagFace embedding $l_{M}$ fixed to inject identity information during the image reconstruction. When analysing the face morph variants, their high visual fidelity and perceived identity consistency qualify them for potential morph attacks. Still, we evaluate whether each face morph variant is equally threatening or whether single samples are particularly security-concerning.

\begin{figure}
\centering
\includegraphics[width=0.99\linewidth]{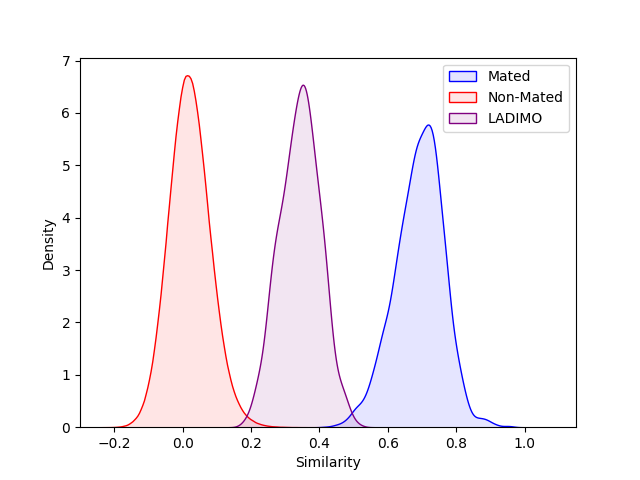}
\caption{Distributions of mated, morph, and non-mated comparison scores, illustrating the elevated attack potential of specific morph variants generated using our proposed re-sampling strategy. Here, all 100 morph comparison scores were computed on a fixed image pair and evaluated on AdaFace~\cite{Kim-AdaFace-CVPR-2022}.}
\label{fig:adaface-single-id-distributions}
\end{figure}

To assess the threat potential of 100 face morph variants with fixed contributing samples, we study morph, mated, and non-mated comparison score distributions in Figure~\ref{fig:adaface-single-id-distributions} computed with AdaFace~\cite{Kim-AdaFace-CVPR-2022}. The wide range of morph comparison scores from 0.2 to almost 0.6 demonstrates the benefit of re-sampling morph variants until one with high attack potential is drawn. In fact, reaching similarity scores of 0.6 causes significant overlaps with the mated comparison score distribution. Thus, to maintain authentication security, the operational threshold must be changed to a level where false reject decisions become more likely.

\begin{figure}[h!]
\centering
\includegraphics[width=0.99\linewidth]{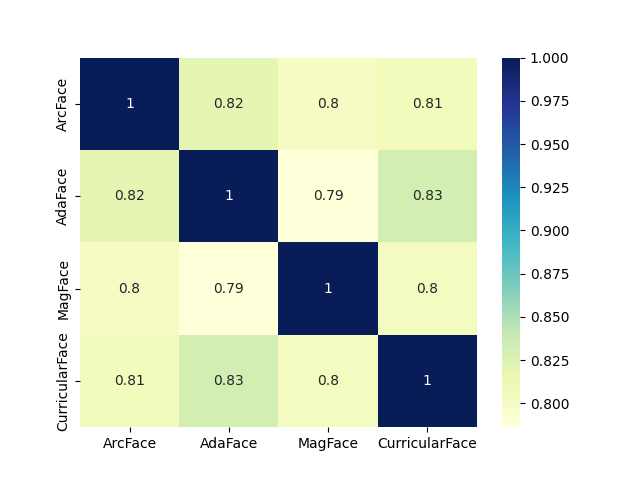}
\caption{Correlation Analysis of the similarity scores obtained by comparing the 100 morph variants to their contributing subjects from the probe dataset across four FRSs \cite{Deng-ArcFace-CVPR-2019}, \cite{Kim-AdaFace-CVPR-2022}, \cite{meng2021magface}, \cite{Huang-Curricularface-CVPR-2020}.}
\label{fig:ladimo-corr-matrix-single-id}
\end{figure}

Ultimately, a face morph variant that threatens a single FRS (like AdaFace in Figure~\ref{fig:adaface-single-id-distributions}), cannot be automatically projected as equally threatening to other FRS. To address this question and evaluate the attack generalizability of high-potential morph variants, we conduct a correlation analysis in Figure~\ref{fig:ladimo-corr-matrix-single-id}. The correlation matrix shows how the face morph comparison scores correlate across four FRS. The high correlation values confirm the high effectiveness of re-sampling morph variants to maximize the attack success rate indifferent to the targeted FRS.    


\section{Summary}
\label{sec:summary}

\subsection{Limitations}

In Figure~\ref{fig:morph-visuals}, LADIMO demonstrates its capability to generate facial images with highly realistic skin textures. However, upon close examination, subtle artefacts, particularly in high-frequency areas like the eye regions, become apparent. Although these artefacts might not be noticeable at first glance, they could potentially influence the success rate of morph attacks when analysed by trained human experts during passport issuance. Despite LADIMO's scalability and ability to be fully automated, the efficiency of diffusion-based image synthesis still lags behind the inference time required by GANs.

\subsection{Future Work}

 The exploration of stochastic morph variations, distinguishing LADIMO from conventional GAN-based~\cite{Zhang-MIPGAN-TBIOM-2021}\cite{Colbois-OptimalFaceMorphs-IJCB-2023} or Diffusion Autoencoder-based\cite{Damer-MorDiff-IWBF-2023}\cite{blasingame2024leveraging} approaches, is encouraged for future work. Treating face morph images as a product of drawing from random distributions has proven to generate face morph variants with diverse ranges of attack potentials. Consequently, an effective strategy to maximize the morph attack potential involves resampling morph variants until obtaining a sample that poses a significant threat across multiple evaluated FRS.

 \subsection{Conclusion}

In conclusion, our proposed Latent Diffusion-based Face Morphing (\textit{LADIMO}) approach demonstrates state-of-the-art performance in the generation of visually convincing face morph attacks with high Morph Attack Potential. By leveraging a Latent Diffusion Model conditioned for biometric template inversion, LADIMO achieves high visual fidelity with realistic skin textures, surpassing the photorealism of GAN-based face morph images. Accordingly, the vulnerability analysis confirms the high attack potential of LADIMO with a high ability to generalize across FRS. The stochastic morph variation analysis underlines the versatility of LADIMO in generating diverse and threatening face morph variants through re-sampling $z_T$. Overall, LADIMO contributes to enriching the diversity in large-scale face morph datasets, supporting the detection of morph attacks and enhancing security in FRS.   

\section*{Acknowledgment}
This research work has been supported by the German Federal Ministry of Education and Research and the Hessian Ministry of Higher Education, Research, Science and the Arts within their joint support of the National Research Center for Applied Cybersecurity ATHENE.

{\small
\bibliographystyle{configs/ieee}
\bibliography{egbib}
}

\end{document}